\def\year{2019}\relax
\documentclass[letterpaper]{article} 
\usepackage{aaai19}  
\usepackage{times}  
\usepackage{helvet}  
\usepackage{courier}  
\usepackage{url}  
\usepackage{graphicx}  
\usepackage{amsmath}
\usepackage{array}
\usepackage{algorithm2e}
\usepackage{algpseudocode,algorithmicx}

\algnewcommand\algorithmicforeach{\textbf{for each}}
\algdef{S}[FOR]{ForEachX}[1]{\algorithmicforeach\ #1\ \algorithmicdo}

\usepackage{subcaption}
\usepackage{multirow}
\usepackage[table,xcdraw]{xcolor}

\newcommand{\ignore}[1]{}
\SetKwInOut{Parameter}{parameter}

\begin{document}

\title{Subset Scanning Over Neural Network Activations}

\author{Skyler Speakman$^1$ \quad Srihari Sridharan$^1$ \quad Sekou Remy$^1$ \quad \\ \\
{\bf \Large Komminist Weldemariam$^1$ \quad Edward McFowland III$^2$} 
       \\
       \\
       $^1$IBM Research \\
       $^2$Carlson School of Management, University of Minnesota }

\maketitle
\begin{abstract}
This work views neural networks as data generating systems and applies anomalous pattern detection techniques on that data in order to detect when a network is processing an anomalous input.  Detecting anomalies is a critical component for multiple machine learning problems including detecting adversarial noise. 
More broadly, this work is a step towards giving neural networks the ability to recognize an out-of-distribution sample.  

This is the first work to introduce ``Subset Scanning'' methods from the anomalous pattern detection domain to the task of detecting anomalous input of neural networks.  Subset scanning treats the detection problem as a search for the ’most anomalous’ subset of node activations (i.e. highest scoring subset according to non-parametric scan statistics). Mathematical properties of these scoring functions allow the search to be completed in log-linear rather than exponential time while still guaranteeing the most anomalous subset of nodes in the network is identified for a given input. 
Quantitative results for detecting and characterizing adversarial noise are provided for CIFAR-10 images on a simple convolutional neural network.
We observe an ``interference'' pattern where anomalous activations in shallow layers suppress the activation structure of the original image in deeper layers.  

\end{abstract}

\section{Introduction}

``Awareness of ignorance is the beginning of wisdom.''

\hspace*{\fill}-- Socrates

We wish to give neural networks the ability to know when they do not know.  In addition, we want networks to be able to explain to a human ``why'' they do not know.  Our approach to this ambitious task is to view neural networks as data generating systems and detect anomalous patterns in the ``activation space'' of their hidden layers.  

\noindent The three goals of applying anomalous pattern detection techniques to data generated by neural networks are to:
\emph{Quantify} the anomalousness of activations within a neural network; 
\emph{Detect} when anomalous patterns are present for a given input; and   
\emph{Characterize} the anomaly by identifying the nodes participating in the anomalous pattern.  

Furthermore, we approach these goals without specialized (re)training techniques or novel network architectures. These methods can be applied to any off-the-shelf, pre-trained model.  
We also emphasize that the adversarial noise detection task is conducted in an unsupervised form, without labeled examples of the noised images.

The primary contribution of this work is to demonstrate that non-parametric scan statistics, efficiently optimized over activations in a neural network, are able to \textbf{quantify} the anomalousness of a high-dimensional input into a real-valued ``score''.  
This definition of anomalousness is with respect to a given network model and a set of ``background'' inputs that are assumed to generate normal or expected patterns in the activation space of the network.
Our novel method measures the deviance between the activations of a given input under evaluation and the activations generated by the background inputs.  A higher measured deviance result 
in a higher anomalousness score for the evaluation input. 

The challenging aspect of measuring deviances in the activation space of neural networks is dealing with high-dimensional data on the order of the number of nodes in a network.  Our baseline example in this work is a convolutional neural network trained on CIFAR-10 images with seven hidden layers and contains 96,800 nodes.
Therefore, the measure of anomalousness must be effective in capturing (potentially subtle) deviances in a high-dimensional space and be computationally tractable.
Subset scanning meets both of these requirements (see Section \ref{subsetscanning}).

The reward for addressing this difficult problem is an unsupervised, anomalous-input detector that can be applied to \emph{any} input and to \emph{any} type of neural network architecture.  This is because neural networks rely on their activation space to encode the features of their inputs and therefore quantifying deviations from expected behavior in the activation space has universal appeal and potential.

We are not analyzing the inputs directly (i.e. the pixel-space) nor performing dimensionality reduction to make the problem more tractable.  We are identifying anomalous patterns at the \emph{node-level} of networks by scanning over subsets of activations and \emph{quantifying} their anomalousness.

The next contributions of this work focus on \textbf{detection} and \textbf{characterization} of adversarial noise added to inputs in order to change the labels \cite{szegedy2013,goodfellow2014fgsm,papernot2016distill}. 

We do not claim state of the art results and do not compare against the numerous and varied approaches in the expanding literature.  
Rather, these results demonstrate that the ``subset score'' of anomalous activations within a neural network is able to \emph{detect} the presence of subtle patterns in high dimensional space.  Also note that a proper ``adversarial noise defense'' is outside the scope of this paper. 

In addition to quantifying the anomalousness of a given input to a network, subset scanning identifies the subset of nodes that contributed to that score.  This data can then be used for \emph{characterizing} the anomalous pattern.  These approaches have broad implications for explainable A.I. by aiding the human interpretability of network models.

For characterizing patterns in this work, we analyze the distribution of nodes (identified as anomalous under the presence of adversarial noise)  across the layers of the network.  We identify an ``interference'' pattern in deeper layers of the network that suggests the structure of activations normally present in clean images have been suppressed, presumably by the anomalous activations from shallower layers.  These types of insights are only possible through the subset scanning approach to anomalous pattern detection.

The final contribution of this work is laying out a line of research that extends subset scanning further into the deep learning domain.  This current paper introduces how to efficiently identify the most anomalous \emph{unconstrained} subset of nodes in a neural network for a single input.  
The subset scanning literature has shown that the unconstrained subset has weak detection power compared to constrained searches where the constraints reflect domain-specific knowledge on the type of anomalous patterns to be detected \cite{neill-ltss-2012,speakman_pfss_2016}.  



The rest of this paper is organized in the following sections.  
Section \ref{subsetscanning} reviews subset scanning and highlights the Linear Time Subset Scanning property originally introduced in \cite{neill-ltss-2012}.  
The section goes on to introduce our novel method that combines subset scanning techniques and non-parametric scan statistics in order to detect anomalous patterns in neural network activations.  
Detection experiments are covered in Section \ref{exp}.  We provide quantitative detection and characterization results on adversarial noise applied to CIFAR-10 images.  Future methodological extensions and new domains of application are covered in Section \ref{ext} and finally, Section \ref{conclusion} provides a summary of the contributions and insights of this work.

\section{Subset Scanning}
\label{subsetscanning}

Subset scanning treats pattern detection as a search for the ``most anomalous'' subset of observations in the data where anomalousness is quantified by a scoring function, $F(S)$ (typically a log-likelihood ratio).  Therefore, we wish to efficiently identify  $S^{\ast} =\arg\max_{S}F(S)$ over all subsets of the data $S$.  The particular scoring functions $F(S)$ used in this work are covered in the next sub-section.  

Subset scanning has been shown to succeed where other heuristic approaches may fail \cite{neill-ltss-2012}.  ``Top-down'' methods look for globally interesting patterns and then identifies sub-partitions to find smaller anomalous groups of records.  These approaches may fail when the true anomaly is not evident from global aggregates.  

Similarly, ``Bottom-up'' methods look for individually anomalous data points and attempt to aggregate them into clusters. These methods may fail when the pattern is only evident by evaluating a group of data points collectively.

Treating the detection problem as a subset scan has desirable statistical properties for maximizing detection power but the exhaustive search is infeasible for even moderately sized data sets.  
However, a large class of scoring functions satisfy the ``Linear Time Subset Scanning'' (LTSS) property which allows for exact, efficient maximization over all subsets of data without requiring an exhaustive search \cite{neill-ltss-2012}.  The following sub-sections highlight a class of functions that satisfy LTSS and describe how the efficient maximization process works for scanning over activations from nodes in a neural network.

\subsection{Nonparametric Scan Statistics}

Subset scanning and scan statistics more broadly 
consider scoring functions that are members of the exponential family and make explicit parametric assumptions on the data generating process.  To avoid these assumptions, this work uses nonparametric scan statistics (NPSS) that have been used in other pattern detection methods \cite{neill-npss-2007,mcfowland-fgss-2013,mcfowland-tess-2018,feng-npss_graph-2014}. 

NPSS require baseline or background data to inform their data distribution under the null hypothesis $H_0$ of no anomaly present.
The evaluation input (different than the background inputs) computes empirical $p$-values by comparing it to the empirical baseline distribution.  NPSS then searches for subsets of data $S$ in the evaluation input that contain the most evidence for \emph{not} having been generated under $H_0$.  This evidence is  quantified by an unexpectedly large number of low empirical $p$-values generated by the evaluation input.


In our specific context, the baseline data $B_{H_0}$ 
is the node activations of 9000 clean CIFAR-10 images from the validation set. Each background image, $X_i$, generates an activation $A^{H_0}_{ij}$ at each network node $j$; and likewise, an evaluation image $X_z$ (which is potentially contaminated with adversarial noise) will produce activations $A_{zj}$.  

For a given evaluation image $X_z$, a collection of background images $X_i \in B_{H_0}$, and a network with $J$ nodes, we can obtain an empirical $p$-value for each node $j$.  This is the proportion of activations from the background inputs, $B_{H_0}$, that are larger than the activation from the evaluation input at node $j$.   
We extend this notion to $p$-value \emph{ranges} which posses improved statistical properties ~\cite{mcfowland-fgss-2013}. 


We use the following two terms to form a $p$-value range at node $j$.

\begin{equation*}
N_{beat}(j) = \sum_{X_i \in B_{H_0}} I(A_{ij} > A_{zj} ) 
\end{equation*}

\begin{equation*}
N_{tie}(j) = \sum_{X_i \in B_{H_0}} I(A_{ij} = A_{zj} ) 
\end{equation*}


The range is then defined as 

\begin{align}
p_j &= \{p^{min}_j, p^{max}_j\} \nonumber \\
    &= \{ \frac{N_{beat}(j)}{|B_{H_0}|+1}, \frac{N_{beat}(j) + N_{tie}(j) + 1}{|B_{H_0}|+1}\}
\end{align}

The empirical $p$-value for node $j$ may then be viewed as random variable uniformly distributed between $p^{min}_j$ and $ p^{max}_j$ under $H_0$.~\cite{mcfowland-tess-2018}.




As an example, consider a node $j$ with activations $A^{H_0}_{ij}$ = $\{-1, -1, 0.2 ,0.75\}$ for $i = 0, 1, 2, 3$ background images.
 An evaluation image $X_z$ may create an activation at node $j$ of $A_{zj} = 0.8$ and would be given a $p$-value range $p_{zj} = (0/5, 1/5)$ for node $j$.  
 A different evaluation image may activate node $j$ at $A_{yj} = 0.1$ and would be given a $p$-value range of $p_{yj} = (2/5, 3/5)$. Finally, a third evaluation image may produce an activation of$A_{xj} = -1$ at node $j$ and would be assigned a range of $p_{xj} = (2/5, 5/5)$.  

Intuitively, if an evaluation image is ``normal'' (its activations are drawn from the same distribution as the baseline images) then few $p$-value ranges will be extreme.  The key assumption for subset scanning approaches is that under the alternative hypothesis of an anomaly present in the data then at least some subset of the activations will appear extreme.  


The $p$-value ranges from an evaluation input are processed by a nonparametric scan statistic in order to identify the subset of node activations that maximizes the scoring function $\max_{S\subseteq P_z}F(S)$, as this is the subset with the most statistical evidence for having been effected by an anomalous pattern. 


The general form of the NPSS score function is 
\begin{equation}
F(S)=\max_{\alpha}F_{\alpha}(S)=\max_{\alpha}\phi(\alpha,N_{\alpha}(S),N(S))
\end{equation}
where $N(S)$ represents the number of empirical $p$-value ranges contained in subset $S$ and $N_{\alpha}(S)$ is the total probability mass less than (significance  level) $\alpha$ in these ranges. 

The $\alpha$ level defines a threshold which $p$-value ranges can be compared against.  Specifically, we calculate the portion of the range that falls below the threshold.  This may be viewed as the probability that a $p$-value from that range would be significant at that $\alpha$ level and is defined as  
\begin{equation}
n_{\alpha}(p_j) = \frac{\alpha - p^{min}_{j}}{p^{max}_{j} - p^{min}_{j}  }
\label{eqn:prio}
\end{equation} 
bounded between 0 and 1. 

This generalizes to a subset of nodes, intuitively.  
$
N_{\alpha}(S) = \sum_{j \in S} n_{\alpha}(p_j)
$
and $N(S) = \sum_{j \in S} 1$.


Moreover, it has been shown that for a subset $S$ consisting of $N(S)$ empirical $p$-value ranges, $E\left[N_{\alpha}(S)\right] = N(S)\alpha$ ~\cite{mcfowland-fgss-2013}. Therefore, we assume an anomalous process will result in some $S$ where the observed significance is higher than the expected, $ N_{\alpha}(S) > N(S)\alpha $, for some $\alpha$.

There are well-known goodness-of-fit statistics that can be utilized in NPSS~\cite{mcfowland-tess-2018}, the most popular is the Kolmogorov-Smirnov test~\cite{kolmogorov-gof-1933}.  Another option is Higher-Criticism ~\cite{donoho-gof-2004}.

In this work we use the Berk-Jones test statistic\cite{berk-bj-1979}: $\phi_{BJ}(\alpha,N_\alpha,N) = N*{KL} \left(\frac{N_\alpha}{N},\alpha\right)$, where $KL$ is the Kullback-Liebler divergence $KL(x,y) = x \log \frac{x}{y} + (1-x) \log \frac{1-x}{1-y}$ between the observed and expected proportions of significant $p$-values. Berk-Jones can be interpreted as the log-likelihood ratio for testing whether the $p$-values are uniformly distributed on $[0,1]$ as compared to following a piece-wise constant alternative distribution, and has been shown to fulfill several optimality properties.

\subsection{Efficient Maximization of NPSS}
Although NPSS provides a means to evaluate the anomalousness of a subset of activations for a given input, discovering which of the $2^J$ possible subsets provides the most evidence of an anomalous pattern is computationally infeasible for large $J$. 
However, NPSS has been shown to satisfy the linear-time subset scanning (LTSS) property \cite{neill-ltss-2012}, which allows for efficient and exact maximization of $\max_{S\subseteq P_z}F(S)$.
For a pair of functions $F(S)$ and $G(j)$ representing the score of a given subset $S$ and the ``priority'' of data record $j$ respectively, we have a guarantee that the subset maximizing the score will be one consisting only of the top-$k$ highest priority records, for some $k$ between $1$ and $J$.

For NPSS the priority of a node activation is the proportion of its $p$-value range that is less than $\alpha$ and was introduced in Equation \ref{eqn:prio}.  $G_{\alpha}(j) = n_{\alpha}(j)$.

\begin{figure*}
  \includegraphics[width = \linewidth]{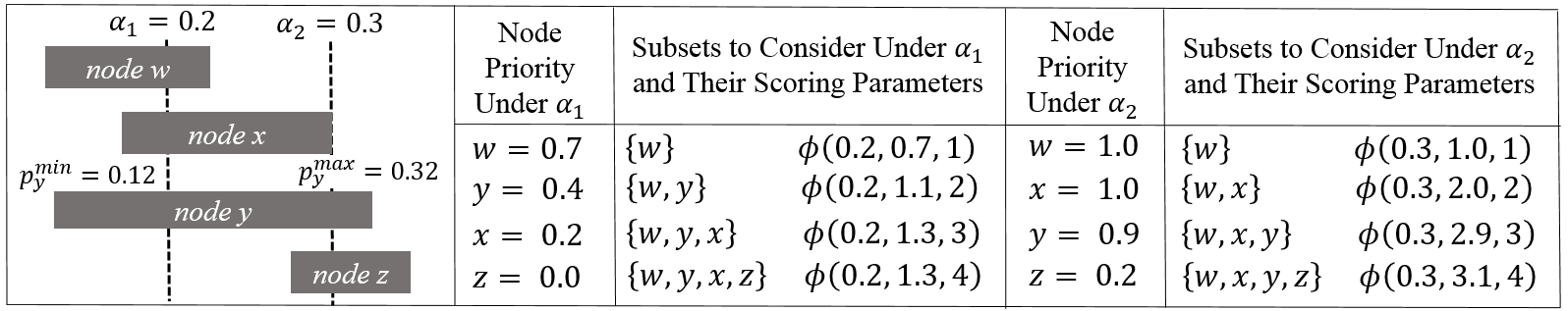}
  \caption{A sample problem with 4 nodes.  The nodes' $p$-value ranges are shown along with two different $\alpha$ thresholds.  The tables show the priority of each node for each threshold and lists the subsets that must be scored in order to guarantee the highest scoring subset is identified.}
  \label{fig:ranges}
\end{figure*}

Figure \ref{fig:ranges} shows a sample problem of maximizing a NPSS scoring function over 4 example nodes and two different $\alpha$ thresholds.  The leftmost graphic shows $p$-value ranges for 4 nodes.  We highlight node $y$ that has $p^{min}_y = 0.12$ and $p^{max}_y = 0.32$.  
For $\alpha = 0.2$ we observe $G_{0.2}(p_y) = 0.4$ as 40\% of node $y$'s $p$-value range is below the threshold. 
A larger proportion of node $w$'s $p$-value range falls below 0.2 and therefore node $w$ has a higher priority than $y$.  

We emphasize that a node's priority (and therefore, also the priority ordering) is induced by the $\alpha$ threshold value.  We demonstrate this by considering $\alpha = 0.3$ in the example, as well.  The priority of node $y$ increases to $G_{0.3}(p_y) = 0.9$.  
Furthermore, the priority \emph{ordering} of the nodes has changed for the different $\alpha$ values.  Node $x$ had a lower priority than node $y$ under $\alpha = 0.2$ and a higher priority than node $y$ under $\alpha = 0.3$.

The next take-away from the example in Figure \ref{fig:ranges} is how the priority ordering over nodes creates at most 4 subsets (linearly many) that must be scored for each $\alpha$ threshold in order to identify the highest-scoring subset overall.  Recall the general form of NPSS scoring functions
$F(S)=\max_{\alpha}F_{\alpha}(S)=\max_{\alpha}\phi(\alpha,N_{\alpha}(S),N(S))$,
where $N(S)$ represents the number of empirical $p$-value ranges contained in subset $S$ and $N_{\alpha}(S)$ is the total probability mass less than (significance  level) $\alpha$ in these ranges.  When scoring the subset $\{w,y\}$ under $\alpha = 0.2$ we evaluate $\phi(0.2, 1.1, 2)$ where 1.1 is the sum of 0.7 and 0.4 and 2 is the size of the subset.  The scoring function is then quantifying how ``anomalous'' it is to observe 1.1 significant $p$-values when the expectation is $(0.2)(2) = 0.4$.

We conclude the toy example by providing intuition behind the efficient maximization of scoring functions that satisfy LTSS. Notice under $\alpha = 0.3$ we do not consider the subset $\{w,x,z\}$.  This is because we can guarantee a higher score by either \emph{including} $y$ or \emph{removing} $z$.  The priority ordering over the nodes guides this inclusion sequence which results in only linearly many subsets needing to be scored.

\subsection{Network Details}

\begin{table}[t]
\caption{Convolutional Neural Network Architecture}
\label{tab:arch}
\setlength{\tabcolsep}{.7em}
\renewcommand{\arraystretch}{1.5}
\centering
\begin{tabular}{|c|c|c|}
\hline
Layer  & Details & \begin{tabular}[c]{@{}c@{}}Number of Nodes\\ in Layer\end{tabular} \\ \hline
Conv 1 & 32, 3x3 & 32,768                                                             \\ \hline
Conv 2 & 32, 3x3 & 28,800                                                             \\ \hline
Pool 1 & 2x2     & 7,200                                                              \\ \hline
\multicolumn{3}{|c|}{Dropout $p = 0.25$}                                              \\ \hline
Conv 3 & 64, 3x3 & 14,400                                                             \\ \hline
Conv 4 & 64, 3x3 & 10,816                                                             \\ \hline
Pool 2 & 2x2     & 2304                                                               \\ \hline
\multicolumn{3}{|c|}{Dropout $p = 0.25$}                                              \\ \hline
Flat   & 512     & 512                                                                \\ \hline
\multicolumn{3}{|c|}{Dropout $p = 0.5$}                                              \\ \hline
\end{tabular}
\end{table}

We briefly describe the training process and network architecture before discussing adversarial attacks.  We trained a standard convolutional neural network on 50,000 CIFAR-10 training images.  The architecture consists of seven hidden layers summarized in Table \ref{tab:arch}.  The first two layers are each composed of 32 3x3 convolution filters.  The third layer is a 2x2 max pooling followed by a dropout of $p = 0.25$.  The next three layers repeat this pattern but with 64 filters in each of the two convolution layers.  Finally there is a flattened layer of 512 nodes with dropout of $p=0.5$ before the output layer.
The model was trained using $tanh$ activation functions and reached a top-1 classification accuracy of 74\%.  The accuracy is within expectation for a simple network. 

$Relu$ activation functions did achieve slightly higher accuracy for the same architecture, however an accurate model is not the focus of this paper and $Relu$ functions can be difficult to identify an ``extreme'' activation due to many of the values being 0 for a given input .  This was evident even if the pre-activation value is anomalously high for the background but still 0 after $Relu$.  
It is possible to perform subset scanning with $relu$ functions with additional constraints.  For example, only allowing positive activations to be considered as part of the most anomalous subset.  These constraints clouded the story and we proceeded with $tanh$ instead.



\section{Detecting Adversarial Noise with Subset Scanning}
\label{exp}
Machine Learning models are susceptible to adversarial perturbations of their input data that can cause the input to be misclassified \cite{szegedy2013,goodfellow2014fgsm,kurakin2016a,dalvi2004}.
There are a variety of methods to make neural networks more robust to adversarial noise.  Some require retraining with altered loss functions so that adversarial images must have a higher perturbation in order to be successful\cite{papernot2015distill,papernot2016distill}.
Other detection methods rely on a supervised approach and treat the problem as classification rather than anomaly detection by training on noised examples \cite{grosse2017,gong2017,huang2015}.  
Another supervised approach is to use activations from hidden layers as features used by the detector. \cite{metzen2017detection}

In contrast, our work treats the problem as anomalous pattern detection and operates in an unsupervised manner without apriori knowledge of the attack or labeled examples.  We also do not rely on training data augmentation or specialized training techniques. These constraints make it a more difficult problem, but also more realistic in the adversarial noise domain as new attacks are constantly being created.  


A defense in \cite{feinman2017detection} is more similar to our work.   They build a kernel density estimate over background activations from the nodes in only the last hidden layer and report when an image falls in a low density part of the density estimate.  This works well on MNIST, but performs poorly on CIFAR-10 \cite{carlini2017bypass}.  Our novel subset scanning approach looks at anomalousness at the node-level and throughout the whole network.

\begin{table}[]
\caption{AUC Detection Results for Subset Scanning over Nodes vs. Scoring All Nodes}
\label{tab:auc}
\setlength{\tabcolsep}{.7em}
\renewcommand{\arraystretch}{1.5}
\centering

\begin{tabular}{|c|c|c|c|}
\hline
\begin{tabular}[c]{@{}c@{}}Attack \\ Type\end{tabular} & \begin{tabular}[c]{@{}c@{}}Attack\\ Parameter\end{tabular} & \begin{tabular}[c]{@{}c@{}}Subset \\ Scanning\end{tabular} & All Nodes \\ \hline

\multicolumn{1}{|l|}{\multirow{3}{*}{FGSM}} & $\epsilon = 0.1$ & 0.9997 & 0.9990 \\ \cline{2-4} 
\multicolumn{1}{|l|}{} & $\epsilon = 0.05$ & 0.9420 & 0.8246 \\ \cline{2-4} 
\multicolumn{1}{|l|}{} & $\epsilon = 0.01$ & 0.5201 & 0.4980 \\ \hline
\multirow{3}{*}{BIM} & $\epsilon = 0.1$ & 0.9913 & 0.9682 \\ \cline{2-4} 
 & $\epsilon = 0.05$ & 0.8755 & 0.6961 \\ \cline{2-4} 
 & $\epsilon = 0.01$ & 0.5177 & 0.4969 \\ \hline
\multirow{3}{*}{CW} & $\kappa = 0$ & 0.5005 & 0.5035 \\ \cline{2-4} 
 & $\kappa = 10$ & 0.5182 & 0.5020 \\ \cline{2-4} 
 & $\kappa = 20$ & 0.5970 & 0.5230 \\ \hline
\end{tabular}
\end{table}

\subsection{Training and Experiment Setup}
For our adversarial experiments, we trained the network described in Section~\ref{subsetscanning} on 50,000 CIFAR-10 images.  We then took $|B_{H_0}| = 9000$ of the 10000 validation images and used them to generate the background activation distribution ($B_{H_0}$) at each of the 96,800 nodes in the network.
The remaining 1000 images were used to form two groups: ``Clean'' (C) and ``Adversarial'' (A) with Adversarial being a noised version for various attack types.  Group C did not change for each attack type.  
We then score the 2000 images contained in A and C.  We emphasize their measure of anomalousness is not between the A and C noised counterparts but rather to the background originally formed by $B_{H_0}$, respectively.

We do not calculate a score threshold for which any input above that score is classified as noise.  Rather we report the area under the ROC curve which is a measure of how well the score separates the classes A and C.  A value of 1.0 means the score perfectly separates the classes and a value of 0.5 is equivalent to random guessing.  

\subsection{Results and Discussion}
Table \ref{tab:auc} provides detection power results (as measured by area under the ROC curve) for a variety of attacks and their parameters.
The three attack types are the Fast Gradient Sign Method (FGSM) \cite{goodfellow2014fgsm} and its iterative extension (BIM) \cite{kurakin2016b}.  These two attacks have an $\epsilon$ parameter which is the maximum distance any pixel in the original image may be changed.  Note this value is in the [0,1] scaled pixel space rather than [0,255].  

The third attack was proposed by Carlini and Wagner (CW) \cite{carlini2017} and has a corresponding $\kappa$ parameter that can create ``high confidence'' attacks.  High confidence is measured by the $\kappa$ difference between the highest and second highest logit values in the output layer (pre-softmax).
All attacks were generated with CleverHans package \cite{papernot2016cleverhans,papernot2018cleverhans}.

All detection results are for the Berk-Jones NPSS scoring function introduced in Section \ref{subsetscanning}.


The first numeric column 
shows the detection power when subset scanning is used to identify the most anomalous subset of activations for the input under evaluation.  The last column shows the detection power when \emph{all} nodes are considered together rather than the highest scoring subset.  Detection power is higher when scanning over subsets of node and demonstrates early promise for expanding subset scanning methods in future work.  

Although still overall low detection power, we point out that our method has a higher probability of detecting the ``higher confidence $\kappa = 20$'' CW attacks than the less confident versions.  This is because the higher confidence attacks require more deviations in the activation space than their lower confidence versions.     

Some attack types did not have 100\% success rate.
BIM failed to change the predicted label on 0.6\% of images for $\epsilon$ of 0.01.
FGSM failed to change the predicted label on 6.1\% ,9.8\%, and 18.8\% of images for $\epsilon =$ 0.10, 0.05, 0.01, respectively. 
CW failed to to generate ``high confidence'' attacks for 17.3\% for $\kappa = 20$.  In all of these cases the failed attacks were removed before calculating detection power.  

\begin{figure}
  \includegraphics[width = \linewidth]{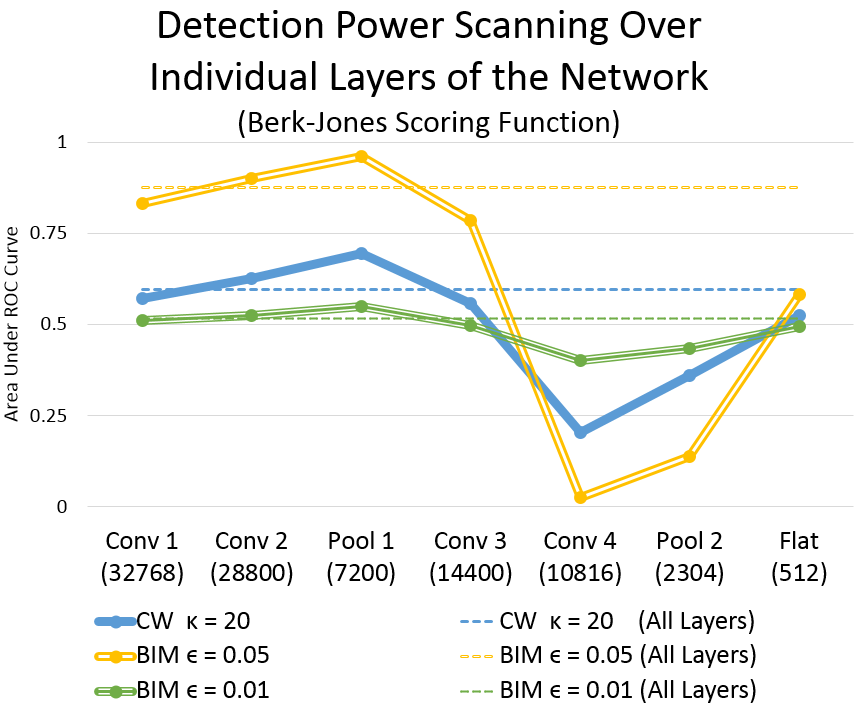}
  \caption{Area under the ROC curve results for scanning over seven individual layers of the network for multiple attacks.  The numbers in parentheses under each layer name is the number of nodes in that particular layer.}
  \label{fig:ind}
\end{figure}

In addition to subset scanning over the entire network we also performed separate searches over individual layers of the network.  This may be thought of as a rudimentary constraint put on the search process, requiring subsets of nodes to be contained in a single layer.  More sophisticated constraints are proposed in detail in Section \ref{ext}.

Figure \ref{fig:ind} shows the detection power of the Berk-Jones scoring function when scanning over \emph{individual} layers of the network. 
We make two observations on these results of increasing importance.  
The first is the increase in detection power when scanning over just the first pooling layer compared to scanning over subsets of nodes in the entire network.
Changes to the pixel space are best captured by the pooling layer that condenses the first two convolution layers.

Second, we note the unexpected behavior at Layer Conv 4 and partly Pool 2.  The AUC values less than 0.5 are due to the score of the most anomalous subset of nodes in Conv 4 from noised images being \emph{less than} the score of the most anomalous subset of nodes in Conv 4 for clean images.  In other words normal activity is \emph{anomalously absent} in those layers for noised images.  
We hypothesize that adversarial noise may more easily confuse neural networks by \emph{de-}constructing the signal of the original image rather than overpowering it with a rogue signal.
This ``interference'' approach results in large amounts of non-interesting activations in the presence of noise compared to the structure of activations for clean images.  Further work is needed on different network architectures to explore this phenomenon.

We conclude the adversarial noise results by locating \emph{where} (i.e. which layer) in the network the most anomalous activations are triggering.  For this approach, we return to subset scanning over the entire network and define a representation metric for each subset $S$ and Layer $L_k$ of the network  $Rep(S,L_k) = \frac{ |S \cap L_k|}{|S|}  /  \frac{|L_k|}{\sum_l|L_l|}$.  
Representation of a subset and layer has a value of 1 if the proportion of anomalous nodes in the subset and the layer is proportional to the relative size of the layer within the network.  This metric allows measuring the relative ``size'' of the subset within a single layer despite layers varying in the number of nodes.  

\begin{figure}[h]
\centering
  \includegraphics[width = 0.75\linewidth]{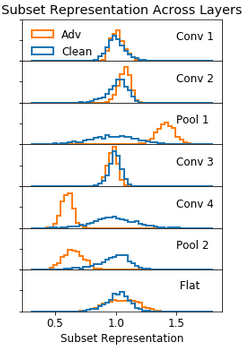}
  \caption{Representation measures the size of the subset in a given layer proportional to the size of the layer in the entire network.  
  Values are shown for the BIM $\epsilon = 0.05$ attack. 
  }
  \label{fig:rep}
\end{figure}

Figure \ref{fig:rep} plots the representation for each subset of the 1000 noised (BIM $\epsilon = 0.05$) and clean images.

We again make two observations of increasing importance.
First, we see that anomalous activity (as identified by subset scanning) of clean images is equally represented across all layers with most subsets having representation centered over 1.0.  

Second, and more consequential,  adversarial images have anomalous activity over-represented in Pool 1 and under-represented in Conv 4 and Pool 2.  
This characterization of anomalous activity, as identified by our method, also suggests the ``interference'' theory of adversarial noise: Anomalous activations in the shallower layers of the network suppress the activation structure of the original image in deeper layers.  

\section{Extensions}
\label{ext}
To enable further 
clarity and applicability of the current work, many extensions of the method have been left for future work.  These extensions may increase detection power or characterization or both. 

\subsection{Simple Extensions}

We note that 2-tailed testing is a reasonable approach for \emph{tanh} and \emph{sigmoid} activation functions.  The definition of ``extreme'' activations carries over to either larger or smaller than expected intuitively.  It is also possible to calculate \emph{density-based} $p$-values where anomalousness is measured by activations from a low density area of the background activations.  This is particularly relevant in deeper nodes where bimodal distributions are likely.  This extension requires learning a univariate kernel density estimate at each node, but this can be done offline on background data only.  

Additionally, we note that it may be worth calculating \emph{conditional} $p$-values where every label has its own set of background activations.  Then at the time of evaluation only the predicted class's background is used for calculating $p$-value ranges.  This may be particularly powerful for the adversarial noise setting, but it does reduce the size of the background activations by the number of classes.

Finally, the NPSS scoring functions have an additional tuning parameter $\alpha_{max}$ that has been left at 1.0 for this work. This means we were able to identify very large subsets of the activations that were all slightly anomalous.  Smaller values of $\alpha_{max}$ limit the search space to a smaller number of more anomalous activations within the network.  Smaller values have been shown to increase detection power further if the prior belief is a small fraction of the data records are participating in the pattern \cite{mcfowland-fgss-2013}.  

\subsection{Enforcing Hard Constraints}
Constraints on the search space are essential parts of subset scanning.  Without constraints, it is likely that inputs drawn from the \emph{null} distribution will look anomalous by chance by ``matching to the noise''.  This hurts detection power despite there being a clear anomalous pattern in the alternative.  
In short, scanning over all subsets may be computationally tractable for scoring functions satisfying LTSS, but it is likely too broad of a search space to capture statistical significance.

This work briefly demonstrated one hard constraint on the search space by performing scans on individual layers of the network as shown in Figure~\ref{fig:ind}.  This simple extension increased detection power when scanning over the first pooling layer compared to scanning over all subsets of the network.  Furthermore, on results not shown in this paper, we are able to increase the detection of CW $\kappa = 20$ from 0.5978 to 0.6553 by scanning over the first 3 layers combined.  

Hard connectivity constraints on subset scanning have been used to identify an anomalous subset of data records that are connected in an underlying graph structure that is either known \cite{speakman-graphscan-2015,speakman_water_tracking} or unknown  \cite{somanchi_graph_structure_learning}. Unfortunately, identifying the highest-scoring connected subset is exponential in the number of nodes and a heuristic alternative could be used to identify a high-scoring 
connected subset \cite{feng-npss_graph-2014}.

\subsection{Enforcing Soft Constraints}
In addition to changing the search space, we may also alter the definition of anomalousness by making changes to the scoring function itself.
For example, we may wish to increase the score of a subset that contains nodes in Pool 1 Layer while decreasing the score of a subset that contains nodes in the Conv 4 Layer.  
These additional terms can be interpreted as the prior log-odds that a given node will be included in the most anomalous subset \cite{speakman_pfss_2016}.

\subsection{Adversarial Noise and Additional Domains}
We emphasize that this work is not proposing a proper defense to adversarial noise.  However, detection is a critical component of a strong defense.  Additional work is needed to turn detection into robustness by leveraging the most anomalous subset activations.

Continuing in the unsupervised fashion, we could mask activations from nodes in certain layers that were deemed anomalous to prevent them from propagating the anomalous pattern.  In a supervised setting, the information contained in the most anomalous subset could be used as features for training a separate classifier.  For example, systematically counting the number of nodes in the most anomalous subset in Pool 1 and Conv 4 could be powerful features.

We note potential for using the subset score to formulate an attack, rather than a tool for detection.  Incorporating the subset score into the loss function of an iterative attack would minimize both the perturbation to the pixel space as well as deviations in the activation space \cite{carlini2017bypass}.

Continuing in the security space, we can also apply subset scanning to data poisoning   \cite{biggio2012poisoning,biggio2013poisoning}.  This current work has considered each image individually, but it is possible to expand it so that the method identifies a \emph{group} of images that are all anomalous for the same reasons in the activation space.  This is the original intention of the Fast Generalized Subset Scan\cite{mcfowland-fgss-2013}.

Leaving the security domain, anomalous pattern detection on neural network activations can be expanded to more general settings of detecting out-of-distribution samples.  This view has implications for detecting bias in classifiers, distribution shift for temporal data, and identifying when new class labels may appear in life-long learning domain.

Finally, we acknowledge that subset scanning over activations of neural networks may have uses in capturing patterns in normal, non-anomalous data.  Identifying which subset of nodes activate higher than expected in a given network while processing normal inputs has implications for explainable A.I. \cite{olah2018the}.

\section{Conclusion}
\label{conclusion}
This work uses the Adversarial Noise domain as an effective narrative device to demonstrate that anomalous patterns in the activation space of neural networks can be \emph{Quantified, Detected, and Characterized.}

The primary contribution of this work to the deep learning literature is a novel, unsupervised anomaly detector that can be applied to any pre-trained, off-the-shelf neural network model.  The method is based on subset scanning which treats the detection problem as a search for the highest scoring (most anomalous) subset of node activations as measured by non-parametric scan statistics.  These scoring functions satisfy the Linear Time Subset Scanning property which allows for exact, efficient maximization over all $2^J$ possible subsets of nodes in a network containing $J$ nodes.

Our method is able to \emph{quantify} activation data on the order of 100,000 dimensions into a single real-valued anomalousness ``score''.  We then used this score to \emph{detect} images that had been perturbed by an adversary in order to change the network's class label of the input.  Finally, we used the identified subset of anomalous nodes in the network to \emph{characterize} the adversarial noise pattern.  This analysis highlighted a possible ``interference'' mode of adversarial noise that uses anomalous activations in the shallow layers to suppress the the true activation pattern of the original image.

We concluded the work by highlighting multiple extensions of subset scanning into the deep learning space. Many of these extensions attempt to overcome the relative weak detection power of \emph{unconstrained} subset scanning that was introduced in this work.  This is accomplished by enforcing constraints on the search space or alterations to the scoring functions, or both.   

Additional domains outside of adversarial noise and security will also benefit from identifying anomalous activity within neural networks.  Life-long learning models need to recognize when a new class of inputs become available and production level systems must always guard against distribution shift over time.  

\newpage

\bibliography{advnoise,tess.bib}
\bibliographystyle{aaai}
\end{document}